%% file: main.tex
\documentclass[11pt,a4paper]{article}
\usepackage[hyperref]{acl2021}
\usepackage{times}
\usepackage{latexsym}

\usepackage[utf8]{inputenc} % allow utf-8 input
\usepackage[T1]{fontenc}    % use 8-bit T1 fonts
\usepackage{hyperref}       % hyperlinks
\usepackage{url}            % simple URL typesetting
\usepackage{booktabs}       % professional-quality tables
\usepackage{amsfonts}       % blackboard math symbols
\usepackage{nicefrac}       % compact symbols for 1/2, etc.
\usepackage{microtype}      % microtypography
\usepackage{amsmath,amsbsy,amsthm,amssymb,bm,color,xcolor}

\usepackage{multicol}
\usepackage{multirow}
\usepackage{caption}
\usepackage{subcaption}
\usepackage{wrapfig}
\usepackage{xspace}
\usepackage{paralist}
\usepackage{makecell}
\usepackage{hyperref}
\usepackage{url}
\usepackage{graphicx}
\aclfinalcopy % Uncomment this line for the final submission
\newcommand{\glm}{GLM\xspace}
\newcommand{\method}{GLAT\xspace}

\title{Glancing Transformer for Non-Autoregressive \\Neural Machine Translation}

\author{Lihua Qian$^{1,2}$\thanks{The work was done when the first author was an intern at Bytedance.} \enskip
Hao Zhou$^{2}$ \enskip
Yu Bao$^{3}$ \enskip
Mingxuan Wang$^{2}$
\\
\textbf{
Lin Qiu$^{1}$ \enskip
Weinan Zhang$^{1}$ \enskip
Yong Yu$^{1}$ \enskip
Lei Li$^{2}$}
\\
$^1$ Shanghai Jiao Tong University \enskip  $^2$ ByteDance AI Lab \enskip  $^3$ Nanjing University \enskip  \\
\texttt{\{qianlihua,lqiu,wnzhang,yyu\}@apex.sjtu.edu.cn}\\
\texttt{\{zhouhao.nlp,wangmingxuan.89,lileilab\}@bytedance.com}\\
\texttt{baoy@smail.nju.edu.cn}\\}

\date{}

\begin{document}
\maketitle

\begin{abstract}
\input{000abstract.tex}
\end{abstract}

\section{Introduction}
\label{sec:intro}
\input{010intro.tex}

\section{Probability Models of Machine Translation}
\label{sec:background}
\input{020back.tex}

\section{Glancing Transformer}
\label{sec:method}

\input{030method.tex}

\section{Experiments}
\label{sec:exp}
\input{040exp.tex}
\section{Related Work}
\label{sec:related}
\input{050related.tex}

\section{Conclusion}
\label{sec:conclusion}
\input{060conclusion.tex}

\bibliographystyle{acl_natbib}
\bibliography{anthology,paper}

%\appendix

\end{document}

%% file: 000abstract.tex
Recent work on non-autoregressive neural machine translation (NAT) aims at improving the efficiency by parallel decoding without sacrificing the quality. 
However, existing NAT methods are either inferior to Transformer or require multiple decoding passes, leading to reduced speedup. 
We propose the Glancing Language Model (\glm), a method to learn word interdependency for single-pass parallel generation models. 
With \glm, we develop Glancing Transformer (\method) for machine translation.
With only single-pass parallel decoding, \method is able to generate high-quality translation with 8$\times$-15$\times$ speedup. 
Experiments on multiple WMT language directions show that \method outperforms all previous single pass non-autoregressive methods, and is nearly comparable to Transformer, reducing the gap to 0.25-0.9 BLEU points. 

%% file: 010intro.tex
\begin{figure}[htb]
    \begin{subfigure}[b]{0.235\textwidth}
    \includegraphics[width=0.8\linewidth]{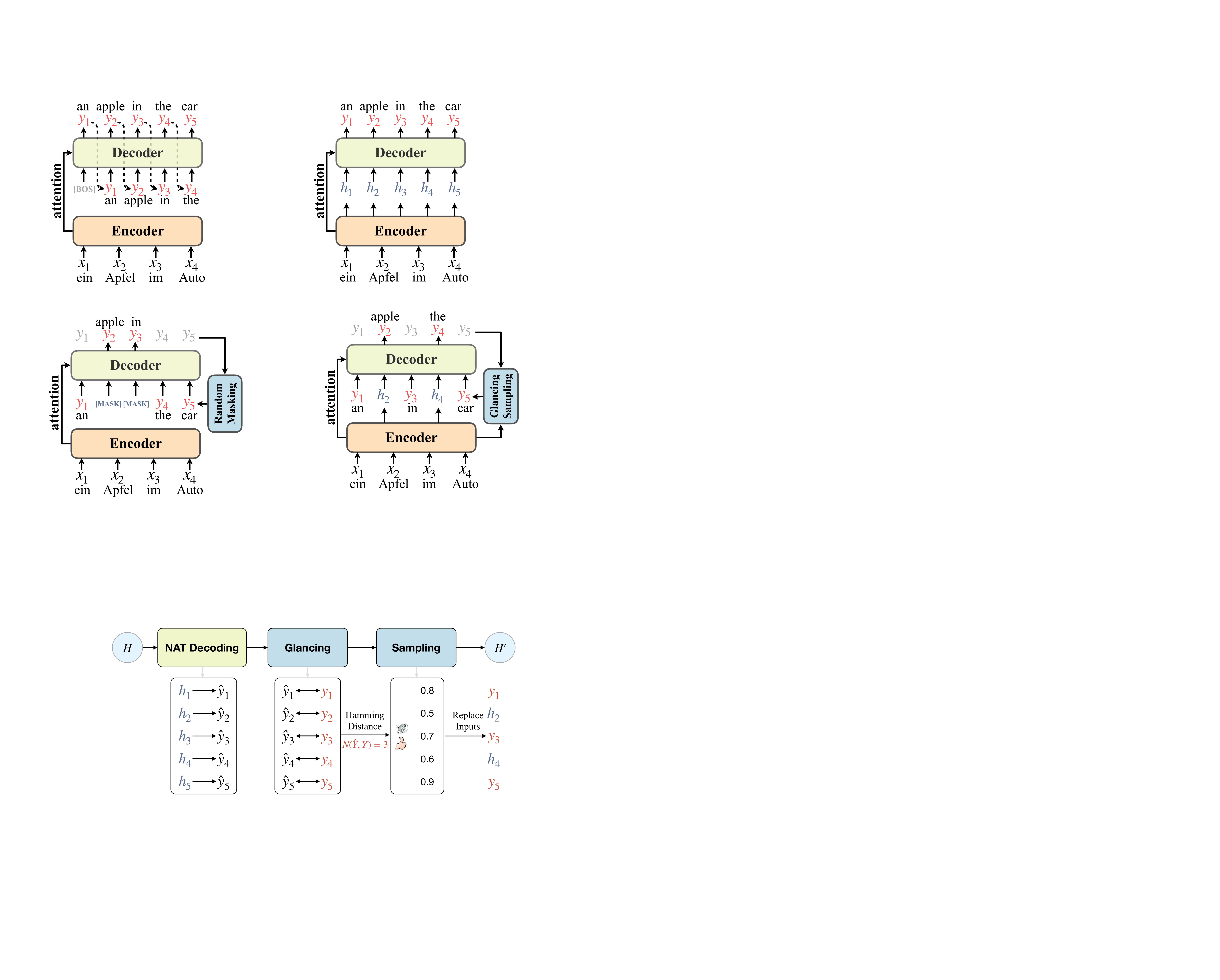}
    \caption{Sequential LM}
    \label{fig:auto_lm}
    \end{subfigure}
    \begin{subfigure}[b]{0.235\textwidth}
    \includegraphics[width=0.8\linewidth]{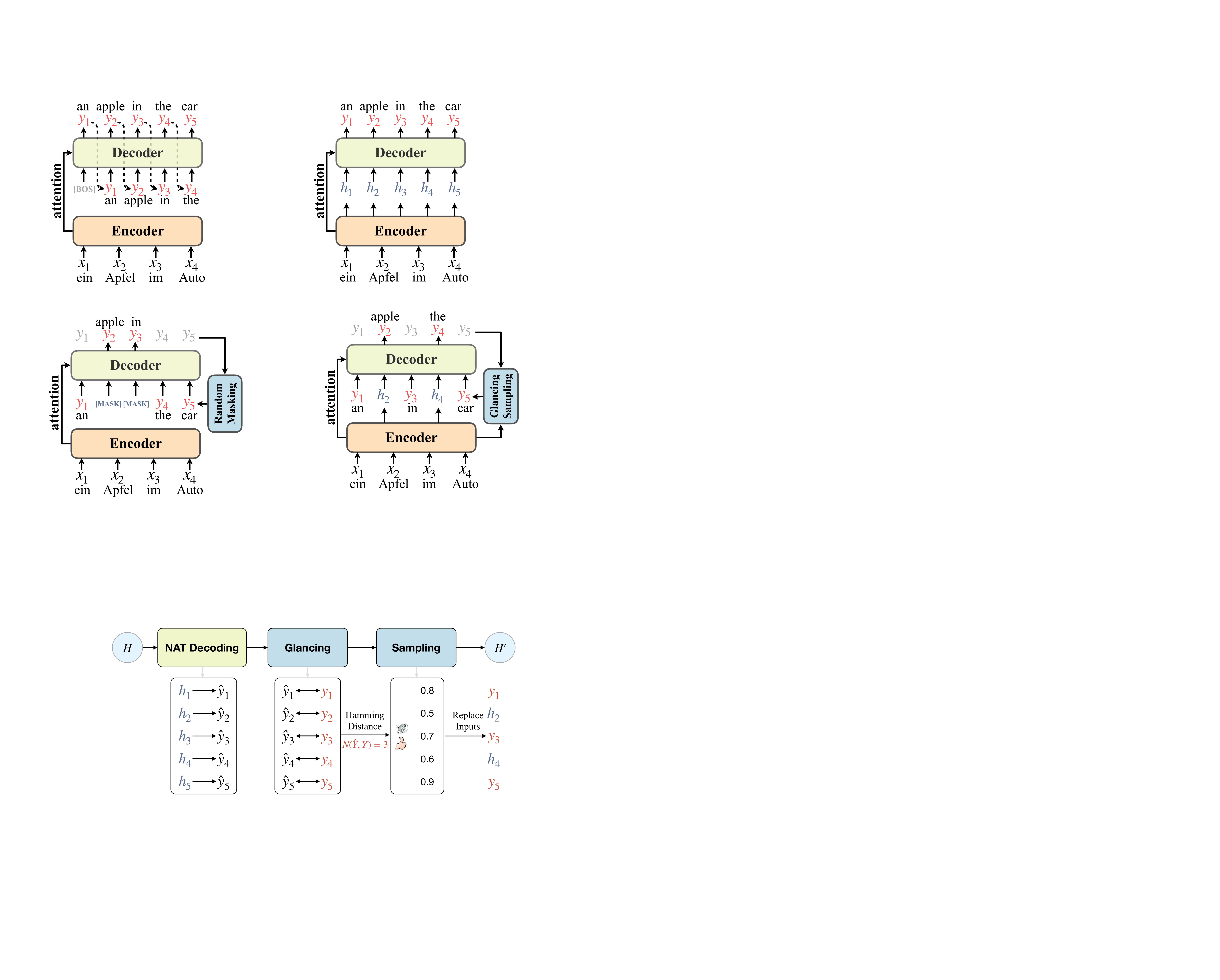}
    \caption{Cond. Independent LM}
    \centering
    \label{fig:nonauto_lm}
    \end{subfigure}
    \begin{subfigure}[b]{0.235\textwidth}
    \includegraphics[width=\linewidth]{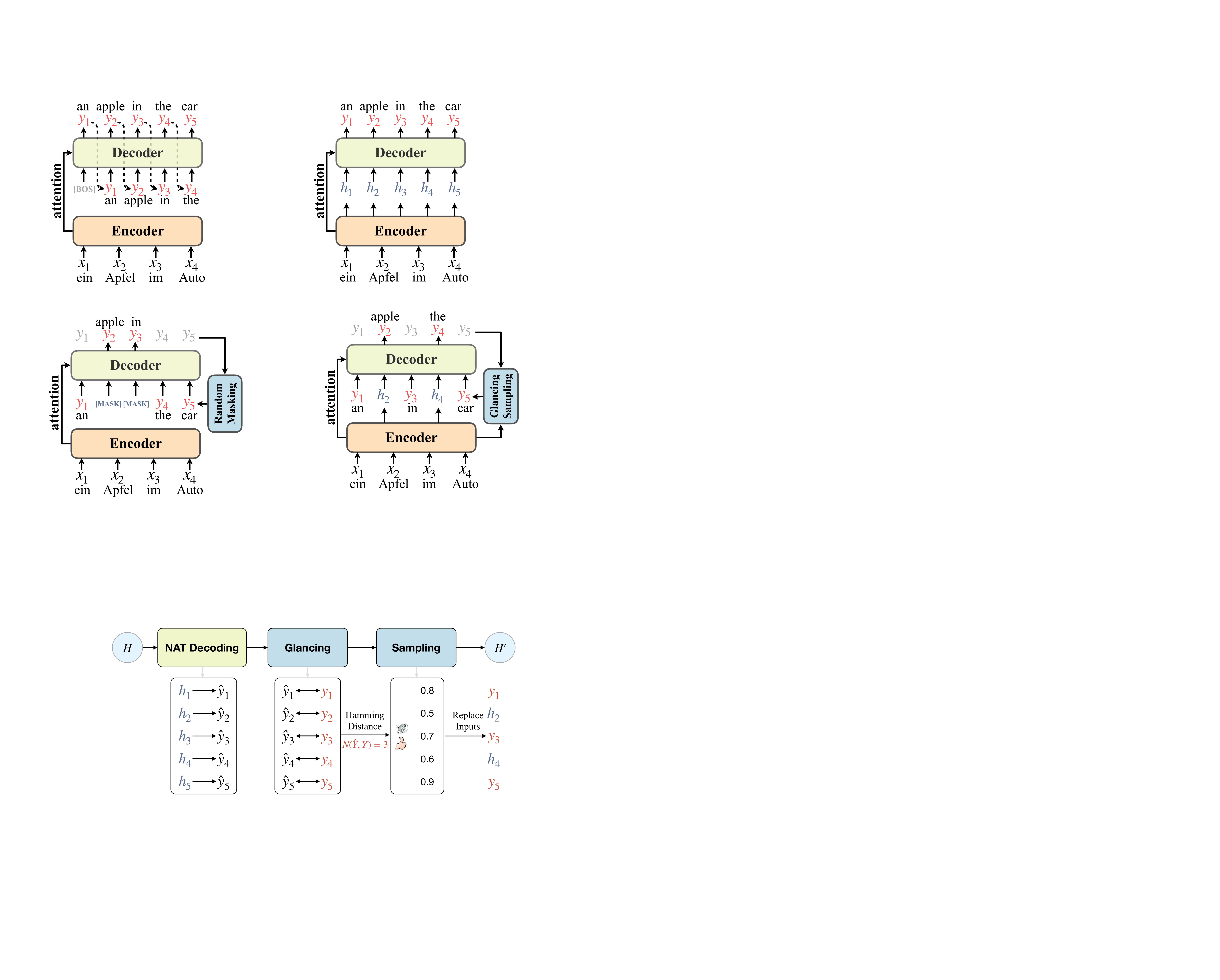}
    \caption{Masked LM (MLM)}
    \label{fig:mlm}
    \end{subfigure}
    \begin{subfigure}[b]{0.235\textwidth}
    \centering
    \includegraphics[width=\linewidth]{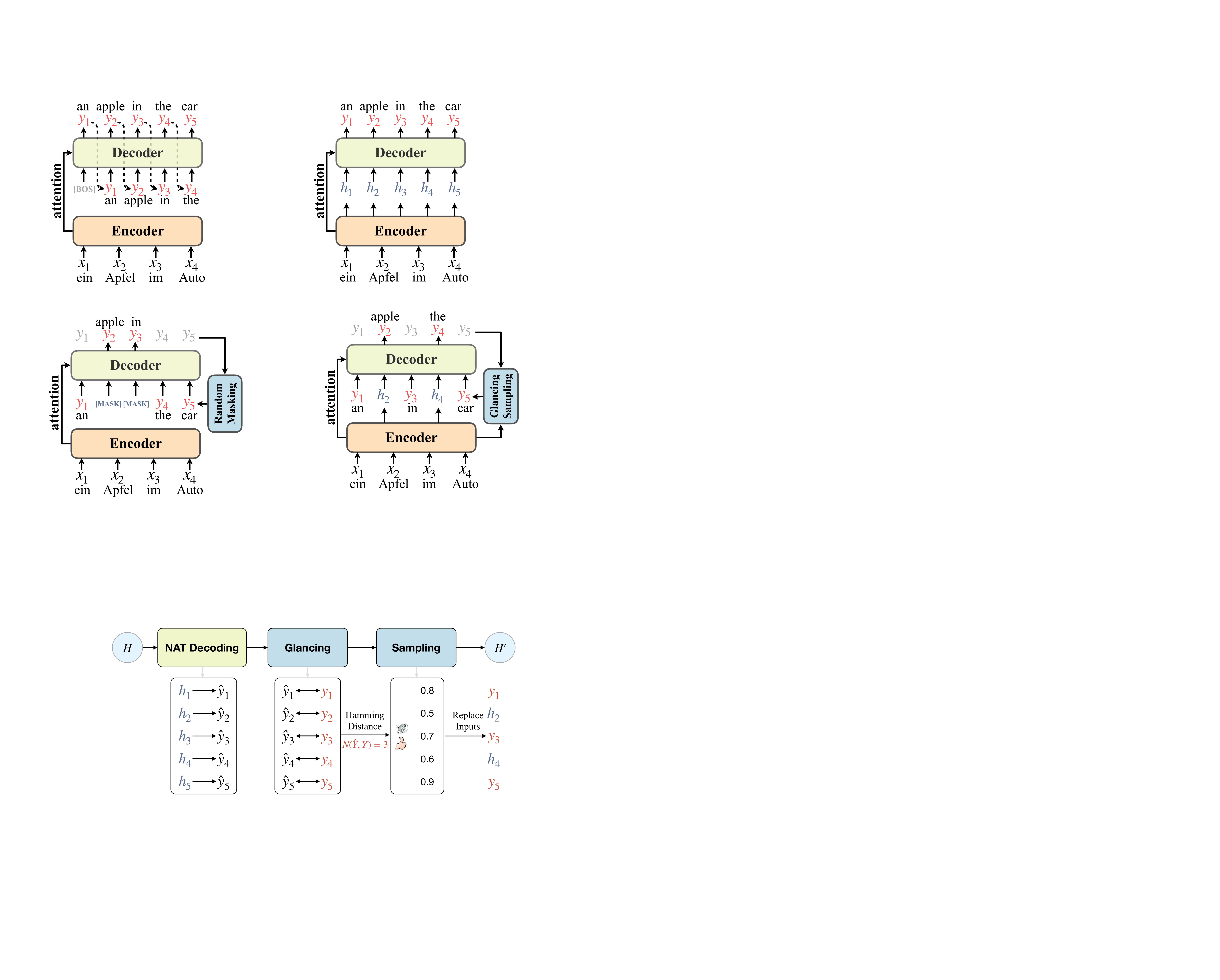}
    \caption{Glancing LM (\glm)}
    \label{fig:glm}
    \end{subfigure}
\caption{Probabilistic models for machine translation methods. 
(\subref{fig:nonauto_lm}) Vanilla NAT uses conditional indepedent LM.
(\subref{fig:mlm}) Mask-Predict NAT uses MLM and requires multiple passes of decoding. (\subref{fig:glm}) Our proposed \glm leverages the decoder prediction to decide glancing sampling policy during training and only requires one pass of decoding during inference. }
\label{fig:lm}
\end{figure}

Transformer has been the most widely used architecture for machine translation~\citep{transformer}.
Despite its strong performance, the decoding of Transformer is inefficient as it adopts the sequential auto-regressive factorization for its probability model (Figure~\ref{fig:auto_lm}). 
Recent work such as non-autoregressive transformer (NAT), aim to decode target tokens in parallel to speed up the generation~\citep{nat}. 
However, the vanilla NAT still lags behind Transformer in the translation quality -- with a gap about 7.0 BLEU score. 
NAT assumes the conditional independence of the target tokens given the source sentence. 
We suspect that NAT's conditional independence assumption prevents learning \emph{word interdependency} in the target sentence. 
Notice that such word interdependency is crucial, as the Transformer explicitly captures that via decoding from left to right (Figure \ref{fig:auto_lm}).

Several remedies are proposed~\citep{mask_predict,levT} to capture word interdependency while keeping parallel decoding.
Their common idea is to decode the target tokens iteratively while each pass of decoding is trained using the \emph{masked language model} (Figure~\ref{fig:mlm}). 
Since these methods require multiple passes of decoding, 
its generation speed is measurably slower than the vanilla NAT. 
With single-pass generation only, these methods still largely lag behind the autoregressive Transformer. %with about 2.0 BLEU points. 

One open question is whether a complete parallel decoding model can achieve comparable machine translation performance to the Transformer. 
It should be non-autoregressive and take only one pass of decoding during the inference time.

% In this paper, we argue that the major culprit of the problem that mask language models have to be used together with multi-pass iterative inference, is the sampling strategy of masking words in MLM. 
% In particular, MLM employs a fixed uniform strategy for randomly masking words during training, which prevents the model from effectively learning word dependencies for one-iteration generation.  
% For example, at the beginning of training when the NAT model is still poorly tuned, we should mask fewer words.
% If we mask too many words, it would be difficult for the NAT model to correctly predict the masked words. 
% On the contrary,  
% if we mask too little words at the end phase of training, the resulting NAT model is rarely trained to predict the whole sentences, and can only predict some sentence fragments. 
% In such a case, to accurately generate the whole sentence in inference, the NAT model has to generate the sentence fragments iteratively. 
% To this end, the sampling strategy is crucial for the training of NAT.

To address the quest, we propose \emph{glancing language model} (GLM), a new method to train a probabilistic sequence model. 
Based on GLM, we develop the \emph{glancing Transformer} (\method) for neural machine translation. 
It achieves parallel text generation with only single decoding pass. 
Yet, it outperforms previous NAT methods and achieves comparable performance as the strong Transformer baseline in multiple cases.
Intuitively, GLM adopts a \textit{adaptive} \textit{glancing sampling} strategy, which glances at some fragments of the reference if the reference is too difficult to fit in the training of GLAT.
Correspondingly, when the model is well tuned, it will adaptively reduce the percentage of glancing sampling, making sure that the resulting model could learn to generate the whole sentence in the single-pass fashion.
% which only improves the training part, while \textit{keeping} the inference part the same as the vanilla NAT model.
% Specifically, \method introduce the .
% Note that our proposed \method 

Specifically, our proposed GLM differs from MLM in two aspects.
Firstly, GLM proposes an adaptive glancing sampling strategy, which enables \method to generate sentences in a one-iteration way, working by gradual training instead of iterative inference~(see Figure \ref{fig:glm}).
% Intuitively, GLM works like the way of 
Generally, GLM is quite similar to curriculum learning~\citep{curriculum} in spirit, namely first learning to generate some fragments and gradually moving to learn the whole sentences~(from easy to hard).
To achieve the adaptive glancing sampling, GLM performs decoding twice in training.
The \textit{first decoding} is the same as the vanilla NAT, and the prediction accuracy indicates whether the current reference is ``difficult'' for fitting.
In the second decoding, GLM gets words of the reference via glancing sampling according to the first decoding, and learn to predict the remaining words that are not sampled. 
Note that only the second decoding will update the model parameters.
% The glancing sampling strategy will sample more words if the predictions for the reference are inaccurate, and the number of sampled words will decrease when \method predicts the reference more accurately along the training process.
% With part of target words as decoder inputs, the target sentence for learning is segmented into small fragments as only remaining target words need to be predicted conditional independently, which smooths the training process and makes the model gradually learn to generate in one iteration.
Secondly, instead of using the \texttt{[MASK]} token, GLM directly uses representations from the encoder at corresponding positions, which is more natural and could enhance the interactions between sampled words and signals from the encoder.

%With part of target words as decoder inputs, the independence assumption is weakened as only the remaining target words are assumed to be conditionally independent.  
%More sampled words will make the learning easier, which smooth the training process appropriately.
%making sure the training of \method in a curriculum learning way. 

% Second, GLM replaces the \texttt{[MASK]} token in the MLM with signals from the encoder, thus not only the target word dependency, but also the dependencies between the input and target are strengthened.
% %This works quite well in practice.
% %\zhouh{add some very short benefits of mixup}
% %GLM smooths the learning curve by start from a easier learning setting that only generate sequence fragments and gradually learn to generate the whole sequence in one-iteration.
% %Besides, GLM strengthens the input representation of decoder by mixing information from both source and target in learning. 
% Equipped with GLM, \method can boost the performance of NAT with one-iteration generation and make multiple decoding iterations no longer necessary.

Experimental results show that \method obtains significant improvements (about 5 BLEU) on standard benchmarks compared to the vanilla NAT, without losing inference speed-up. 
\method achieves competitive results against iterative approaches like Mask-Predict~\citep{mask_predict}, even outperforming the Mask-Predict model on WMT14 DE-EN and WMT16 RO-EN.
% Considering the fully NAT models with the one-iteration simultaneous generation, \method achieves the best BLEU scores.
Compared to the strong AT baseline, \method can still close the performance gap within 0.9 BLEU point while keeping $7.9\times$ speed-up.
Empirically, we even find that \method outperforms AT when the length of the reference is less than 20 on WMT14 DE-EN.
We speculate this is because GLM could capture bidirectional context for generation while its left-to-right counterpart is only unidirectional, which indicates the potential of parallel generation approaches like \method.

%% file: 020back.tex
We state and compare different probability models for machine translation. 
A machine translation task can be formally defined as a sequence to sequence generation problem: given the source sentence $X=\{x_1,x_2,...,x_N\}$, 
to generate the target sentence $Y=\{y_1,y_2,...,y_T\}$
according to the conditional probability $P(Y|X;\theta)$, where $\theta$ denotes the parameter set of a network. 
Different methods factorize the conditional probability differently. 

The Transformer uses the autoregressive factorization to maximize the following likelihood:
\[
\mathcal{L}_{\text{AT}}=\log P(Y|X;\theta)=\sum_{t=1}^{T} \log p(y_t|y_{<t},X;\theta),
\]
where $y_{<t}=\{\texttt{[BOS]},y_1,...,y_{t-1}\}$.
For simplicity, we omit the number of samples in the equation. 
Note the training of AT adopts left-to-right teacher forcing on the target tokens~\citep{transformer}. 
The word interdependency is learned in a unidirectional way.
During inference, the preceding predicted token is fed into the decoder to generate the next token.

The vanilla NAT consists of the same encoder as Transformer and a parallel decoder with layers of multi-head attention~\citep{nat}. 
During training, it uses the conditional independent factorization for the target sentence:
\[
\mathcal{L}_{\text{NAT}}=\sum_{t=1}^{T} \log P(y_t|X;\theta).\\
\]
Notice that, NAT's log-likelihood is an approximation to the full log-likelihood $\log P(Y|X; \theta)$. 
During inference, the encoder representation is copied as the input to the decoder, therefore all tokens on the target side can be generated in parallel. 
Such a conditional independence assumption does not hold in general, which explains the inferior performance of NAT.

\begin{figure*}[htb]
\centering
\includegraphics[width=0.8\linewidth]{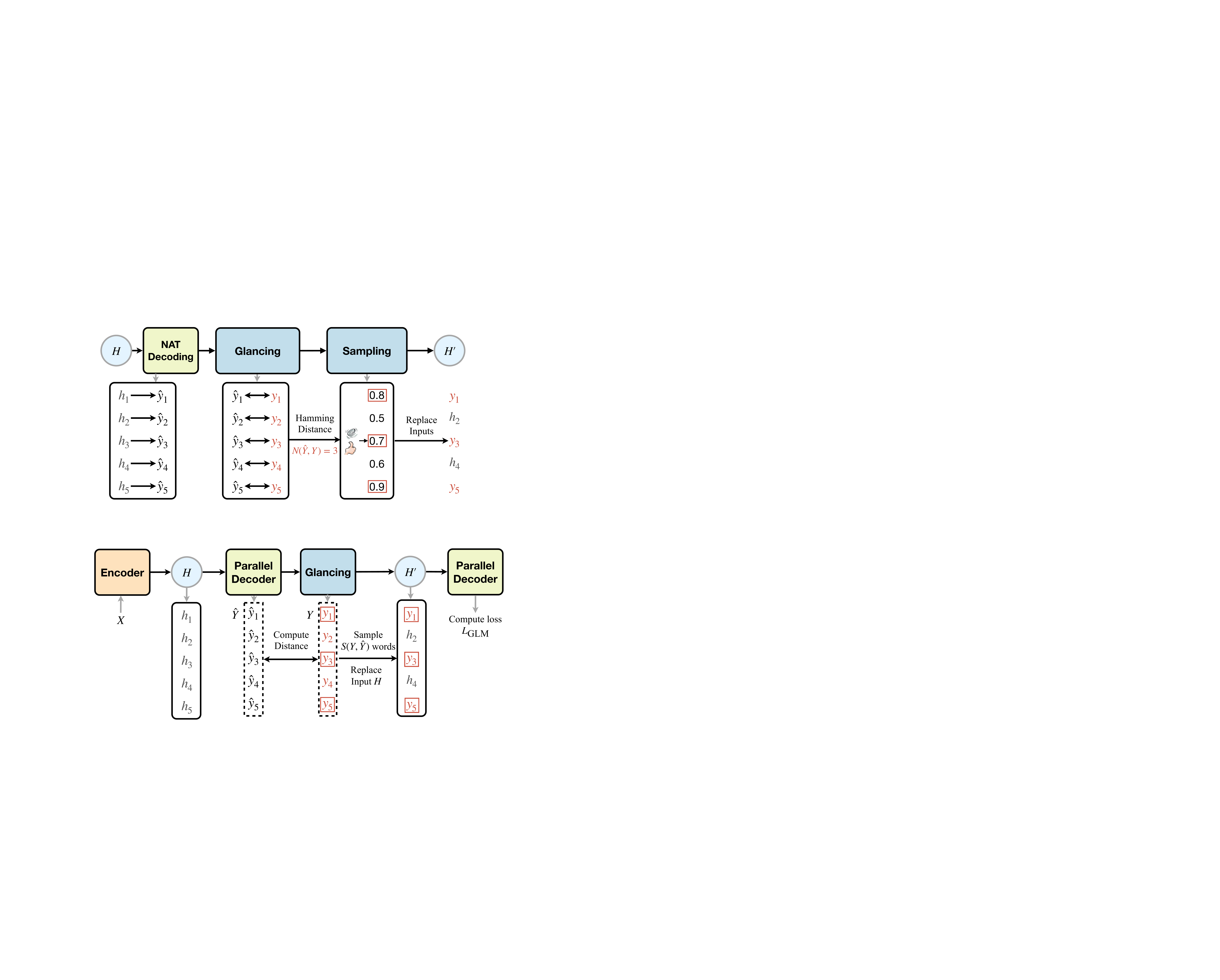}
\caption{The training procedure with glancing sampling in \method. $H$ is the representation computed by the encoder. $\hat{y}$'s are the initial predicted tokens of the parallel decoder. $y$'s are the ground-truth target tokens. $H'$ is fed into the decoder again to calculate the training loss. }
\label{fig:arch}
\end{figure*}

Multi-pass iterative decoding approaches such as Mask-Predict~\citep{mask_predict} extends the vanilla NAT. 
It still uses the conditional independent factorization, together with the random masking scheme:
\[
\mathcal{L}_{\text{MLM}} = \sum_{y_t \in \mathbb{RM}(Y)} \log p\Big(y_t|\Phi\big(Y,\mathbb{RM}(Y)\big),X; \theta\Big),
\]
where $\mathbb{RM}(Y)$ is a set of randomly selected words from $Y$, and $\Phi(\cdot)$ replaces these selected words in $Y$ with the \texttt{[MASK]} token.
For example in Figure~\ref{fig:mlm}, $\mathbb{RM}(Y) = \{y_2, y_3\}$,  $\Phi\big(Y,\mathbb{RM}(Y)\big) = \{y_1, \texttt{[MASK]}, \texttt{[MASK]}, y_4,y_5\}$.   
The training objective is to learn a refinement model $\theta$ that can predict the masked tokens given the source sentence $X$ and words generated in the previous iteration.

The vanilla NAT breaks word interdependency, while MLM requires multiple passes of decoding to re-establish the word interdependency. 
Our goal in this work is to design a better probability model and a training objective to enable word interdependency learning for single-pass parallel generation. 

%% file: 030method.tex
In this section, we present \method in detail.
\method uses the same encoder-decoder architecture as the vanilla NAT~\citep{nat}. 
\method differs from the vanilla NAT in that it explicitly encourages word interdependency via training with glancing language model~(\glm). 
It differs from the iterative NAT with MLM in that it is trained to produce single pass parallel decoding while MLM is used for prediction refinement.

%The GLM in \method gradually learns longer sentence fragments for simultaneous generation and introduces explicit dependency modeling with the mix of source and target inputs, which better captures dependencies under the one-iteration generation setting.
%The \glance technique is designed to overcome the two disadvantages of NAT~(discussed in Section 2), which makes the learning easier by gradually approaching learning under the complete conditional independence assumption and avoids \deadlock by introducing explicit target language modeling.

% Generally, \method performs the two-pass decoding of NAT during training.
% At the first pass, we get the output prediction and compare it with the reference~(ground truth sentence), then sample target words by glancing at the reference according to how well this reference is predicted.
% At the second pass, we feed the sampled words to the decoder at their corresponding positions, then train the model to predict the remaining words via maximum likelihood estimation.
% In our proposed \method, more reference words for inaccurate sentence predictions tend to be sampled as decoding inputs.
% As the model can better predict the reference along the training process, the model glances fewer reference words, thus the target fragments to be learned grows gradually.
% \method also introduces explicit dependency learning in predicting remaining words with some target words as input and strengthens the decoder input with source representation.

\subsection{The Glancing Language Model}
Given the input source sentence $X=\{x_1,x_2,...,x_N\}$, the task is to predict $Y=\{y_1,y_2,...,y_T\}$. 
The glancing Transformer (\method) formulates a glancing language model (\glm) during training. It maximizes the following:
\begin{equation}
% \begin{split}
    \mathcal{L}_{\text{\glm}}= 
    \sum_{y_t \in \overline{\mathbb{GS}(Y,\hat{Y})}}\\
     \log p(y_t| \mathbb{GS}(Y,\hat{Y}), X; \theta)\label{eqn:objective}
% \end{split}
\end{equation}
Where, $\hat{Y}$ is an initial predicted tokens, 
and $\mathbb{GS}(Y,\hat{Y})$ is a subset of tokens selected via the \emph{glancing sampling} strategy (Figure~\ref{fig:arch}, described in detail in the next section).
The glancing sampling strategy selects those words from the target sentence by comparing the initial prediction against the ground-truth tokens. 
It selects more tokens and feeds them into the decoder input if the network's initial prediction is less accurate. 
$\overline{\mathbb{GS}(Y,\hat{Y})}$ is the remaining subset of tokens within the target $Y$ but not selected. 
The training loss above is calculated against these remaining tokens.

% \begin{equation}
% p(Y|X; \theta)=\prod_{t=1}^{T}p(y_t|y_{<t},X; \theta),\\
% \end{equation}
% In the following sections, we will describe the whole learning process of glancing language model and the sampling strategy in detail, respectively.

\method adopts similar encoder-decoder architecture as the Transformer with some modification~(Figure~\ref{fig:glm}). 
Its encoder $f_{\text{enc}}$is the same multi-head attention layers.
Its decoder $f_{\text{dec}}$ include multiple layers of multi-head attention where each layer attends to the full sequence of both encoder representation and the previous layer of decoder representation. 
%It also includes one additional neural layer to predict the length of the output sequence. 

During the initial prediction, the input to the decoder  $H = \{h_1,h_2,...,h_T\}$  are copied from the encoder output using either \emph{uniform copy} or \emph{soft copy}~\citep{imitate_nat}. 
The initial tokens $\hat{Y}$ are predicted using $\operatorname{argmax}$ decoding with
 $f_{\text{dec}}(f_\text{enc}(X;\theta), H; \theta)$.

To calculate the loss $\mathcal{L}_{\text{\glm}}$, we compare the initial prediction $\hat{Y}$ against the ground-truth to select tokens within the target sentence, i.e. $\mathbb{GS}(Y,\hat{Y})$. 
We then replace those sampled indices of $h$'s with corresponding target word embeddings, $H' = \mathbb{RP}(\text{Emb}_{y_t\in \mathbb{GS}(Y,\hat{Y})}(y_t), H)$, where $\mathbb{RP}$ replaces the corresponding indices.
Namely, if a token in the target is sampled,  its word embedding
replaces the corresponding $h$. 
Here the word embeddings are obtained from the softmax embedding matrix of the decoder.
The updated $H'$ is then fed into the decoder $f_{\text{dec}}$ again to calculate the output token probability. Specifically, the output probabilities of remaining tokens $p(y_t|\mathbb{GS}(Y,\hat{Y}), X; \theta)$ are computed with $f_\text{dec}(H', f_\text{enc}(X;\theta);\theta)$.

\subsection{The Glancing Sampling Strategy}
One important component of \glm is to adaptively select the positions of tokens from the target sentence.
Those selected tokens provide ``correct'' information from the ground-truth target, therefore it helps training the decoder to predict the rest non-selected tokens. 
Intuitively, our adaptive sampling strategy guides the model to first learn the generation of fragments and then gradually turn to the whole sentences.
Our glancing sampling strategy selects many words at the start of the training, when the model is not yet well tuned. 
As the model gets better progressively, the sampling strategy will sample fewer words to enable the model to learn the parallel generation of the whole sentence.
Note that the sampling strategy is crucial in the training of \method.

As illustrated in Figure~\ref{fig:arch}, the glancing sampling could be divided into two steps: first deciding a sampling number $S$, and then \textit{randomly} selecting $S$ words from the reference.
The sampling number $S$ will be larger when the model is poorly trained and decreases along the training process.
Note that we choose to randomly select the $S$ words from the reference. The random reference word selection is simple and yields good performance empirically.
%instead of choosing the most uncertain~(with lower prediction scores in the first decoding) or certain words, 
% Many studies enjoys the it always offers simple implementation and desirable performance in practice~\citep{bert}.

Formally, given the input $X$, its predicted sentence $\hat Y$ and its reference $Y$, the goal of glancing sampling function $\mathbb{GS}(Y,\hat{Y})$ is to obtain a subset of words sampled from $Y$:
\begin{equation}
    \mathbb{GS}(Y,\hat{Y}) = \text{Random}(Y, S(Y, \hat{Y}))
\end{equation}
Here, $\text{Random}(Y, S)$ is randomly selecting $S$ tokens from $Y$, and $S$ is computed by comparing the difference between $\hat{Y}$ and $Y$, 
$S(Y, \hat{Y}) = \lambda \cdot d(Y, \hat{Y})$.
The sampling ratio $\lambda$ is a hyper-parameter to more flexibly control the number of sampled tokens. 
$d(Y,\hat Y)$ is a metric for measuring the differences between $Y$ and $\hat Y$. 
We adopt the Hamming distance~\citep{hamming} as the metric, which is computed as $d(Y,\hat Y)=\sum_{t=1}^T (y_t \neq \hat y_t)$.
With $d(Y,\hat Y)$, the sampling number can be decided adaptively considering the current trained model's prediction capability.
% Note that $d(Y,\hat Y)$ could be other distances such as Levenshtein distance~\citep{lev_dis}, but we find the Hamming distance achieves the best result empirically.
For situations that $Y$ and $\hat{Y}$ have different lengths, $d(Y,\hat Y)$ could be other distances such as Levenshtein distance~\citep{lev_dis}.

Alternative glancing sampling strategy can be adopted as well. For example, one simple alternative strategy is to set the number of sampled tokens to be proportional to the target sentence length, i.e. $S=\lambda * T$. We will evaluate the effects of these variations in the experiment.

% \subsection{Knowledge Distillation}
% Following previous work~\citep{nat,iter_nat,nat_reg}, we also use sequence-level knowledge distillation for all datasets. 
% We employ the transformer with base setting in ~\citet{transformer} as the teacher for knowledge distillation. Then, we train our \method on distilled data.

\subsection{Inference}
\method only modifies the training procedure. 
Its inference is fully parallel with only a single pass.
For parallel generation, we need to decide the output lengths before decoding. A simple way to decide the output lengths is predicting length with representations from the encoder.

In \method, the length prediction is implemented as in~\citet{mask_predict}. An additional \texttt{[LENGTH]} token is added to the source input, and the encoder output for the \texttt{[LENGTH]} token is used to predict the length.

We also use two more complex methods to better decide the output lengths: noisy parallel decoding~(NPD) and connectionist temporal classification~(CTC).
For NPD~\citep{nat}, we first predict $m$ target length candidates, then generate output sequences with $\operatorname{argmax}$ decoding for each target length candidate. 
Then we use a pre-trained transformer to rank these sequences and identify the best overall output as the final output.
For CTC~\citep{ctc}, following~\citet{nat_ctc}, we first set the max output length to twice the source input length, and remove the blanks and repeated tokens after generation.
% In order to obtain the best target length, we also consider the common practice of noise parallel decoding~\citep{nat,iter_nat,enat,nat_reg}, which generates several decoding candidates in parallel and selects the best via re-scoring with a pre-trained autoregressive model.
% For~\method, we first predict $m$ target length candidates, then generate output sequences with $\operatorname{argmax}$ decoding for each target length candidate.
%, which is also called length parallel decoding~(LPD)~\citep{imitate_nat}.

%% file: 040exp.tex
\begin{table*}[!tbp]
\centering
\small
\tabcolsep 4pt
 \scalebox{0.9}{
\begin{tabular}{lcl|c|cccc|c}
\toprule
\multicolumn{3}{c|}{\multirow{2}{*}{Models}} & \multirow{2}{*}{$I_{\text{dec}}$} & \multicolumn{2}{c}{WMT14} &\multicolumn{2}{c|}{WMT16} &\multirow{2}{*}{Speed Up} \\
\multicolumn{3}{c|}{} &  & EN-DE & DE-EN & EN-RO & RO-EN &        \\
\midrule
\multicolumn{2}{c}{\multirow{2}{*}{AT Models}}  
  &Transformer~\citep{transformer} & T & 27.30 & / & / & / &  / \\
 \multicolumn{2}{c}{}  & Transformer~(ours)   & T & 27.48 & 31.27 & 33.70 & 34.05 & 1.0$\times$ \\
\midrule
 \multicolumn{2}{c}{\multirow{5}{*}{Iterative NAT}}& NAT-IR~\citep{iter_nat}   & 10   & 21.61 & 25.48 & 29.32 & 30.19 & 1.5$\times$ \\
 \multicolumn{2}{c}{}& LaNMT~\citep{shu2020latent}  & 4     & 26.30 & /     & /     & 29.10 & 5.7$\times$  \\
 \multicolumn{2}{c}{}& LevT~\citep{levT}  & 6+     & 27.27 & /     & /     & 33.26 & 4.0$\times$  \\
 %& Mask-Predict$_\text{small}$~($I_{dec}$=10)&25.51& 29.47& 31.65 & 32.27 & / \\
%  & Mask-Predict & 1 & 18.05 & 21.83 & 27.32 & 28.20 &  /           \\
 \multicolumn{2}{c}{}& Mask-Predict~\citep{mask_predict} & 10 &27.03& 30.53&\textbf{33.08} &
 \textbf{33.31} & 1.7$\times$    \\
 \multicolumn{2}{c}{}& JM-NAT~\citep{jm_nat} & 10 & \textbf{27.31} & \textbf{31.02} & / & / & 5.7$\times$\\
 \midrule
\multirow{18}{*}{\ \ Fully NAT\ \ } &
 & NAT-FT~\citep{nat}               & 1 & 17.69 & 21.47 & 27.29 & 29.06 & 15.6$\times$\\
 %& NAT-IR              & 13.91 & 16.77 & 24.45  & 25.73 &  9.0$\times$
 %\\
%  & NAT-REG              & 1 & 20.65 & 24.77 & /     & /     & / \\%27.6$\times$ \\
 & & Mask-Predict~\citep{mask_predict} & 1 & 18.05 & 21.83 & 27.32 & 28.20 &  /           \\
  & & imit-NAT~\citep{imitate_nat}          & 1 & 22.44 & 25.67 & 28.61 & 28.90 & 18.6$\times$ \\
 & & NAT-HINT~\citep{hint_nat}              & 1 & 21.11 & 25.24 & /     & /     & / \\%30.2$\times$ \\
 %  & Flowseq              & 1 & 21.45 & 26.16 & 29.34 & 30.44 &  /           \\
 & & Flowseq~\citep{flowseq}              & 1 & 23.72 & 28.39 & 29.73 & 30.72 &  1.1$\times$           \\ 
 & & NAT-DCRF~\citep{nat-crf}            & 1 & 23.44 & 27.22 & / & / & 10.4 $\times$ \\
%  & Mask-Predict$_\text{small}$~($I_{dec}$=1)  & 15.06 & 19.26 & 20.12 & 20.36 &  /           \\
%  & Mask-Predict$_\text{base}$ & 1 & 18.05 & 21.83 & 27.32 & 28.20 &  /           \\
 % & Imputer              & 1 & \textbf{25.8} & 28.4 & / & / & / \\
%  & NAT-base~(ours) & 1 & 20.36 & 24.81 & 28.47 & 29.43 &  15.3$\times$           \\
%  & \method~(ours)    & 1    & \textbf{25.21}& \textbf{29.84} & \textbf{31.19} & \textbf{32.04} & 15.3$\times$          \\
%\hline 
 \cmidrule{2-9}
  & \multirow{2}{*}{w/ CTC}
 & NAT-CTC~\citep{nat_ctc} & 1 & 16.56 & 18.64 & 19.54 & 24.67 & / \\
 & & Imputer~\citep{imputer}              & 1 & 25.80 & 28.40 & 32.30 & 31.70 & 18.6$\times$ \\
 
 \cmidrule{2-9}
 %\multicolumn{2}{c}{\multirow{5}{*}{\begin{tabular}[c]{@{}c@{}}Fully NAT \\ w/  NPD\end{tabular}}}
 & \multirow{5}{*}{w/ NPD} 
%  & NAT-FT~(m=10)         & 18.66 & 22.41 & 29.02 & 30.76 & 7.7$\times$  \\
 & NAT-FT + NPD~(m=100)      & 1  & 19.17 & 23.20 & 29.79 & 31.44 & 2.4$\times$  \\
  %& NAT-REG~(m=9)       & 1  & 24.61 & 28.90 & /     & /     & 15.1$\times$ \\
  & & imit-NAT + NPD~(m=7)   & 1  & 24.15 & 27.28 & 31.45 & 31.81 & 9.7$\times$ \\
  & & NAT-HINT + NPD~(m=9)       & 1  & 25.20 & 29.52 & /     & /     & / \\%17.8$\times$ \\
 %  & Flowseq~(m=30)      & 1  & 23.48 & 28.40 & 31.75 & 32.49 & /           \\
  & & Flowseq + NPD~(m=30)      & 1  & 25.31 & 30.68 & 32.20 & 32.84 & /           \\
 & & NAT-DCRF + NPD~(m=9)     & 1  & 26.07 & 29.68 & / & / & 6.1$\times$\\
%  &\method~(NPD m=7, ours)  & 1   & \textbf{26.55} &\textbf{31.02}&\textbf{32.87}&\textbf{33.51} & 7.9$\times$  \\
  
%   & CTC~(ours)              & 1 & 25.52 & 28.73 & / & / & 14.6 $\times$ \\
%   & \method+CTC~(ours)              & 1 & 26.39 & 29.54 & / & / & 14.6 $\times$ \\
 \cmidrule{2-9}
  & \multirow{5}{*}{\textbf{Ours}} 
  & NAT-base* & 1 & 20.36 & 24.81 & 28.47 & 29.43 &  15.3$\times$           \\
  & & CTC*              & 1 & 25.52 & 28.73 & 32.60 & 33.46 & 14.6 $\times$ \\
  & & \method    & 1    & 25.21& 29.84 & 31.19 & 32.04 & 15.3$\times$          \\
 & & \method + CTC            & 1 & 26.39 & 29.54 & 32.79 & \textbf{33.84} & 14.6 $\times$ \\
  & &\method + NPD ~(m=7)  & 1   & \textbf{26.55} &\textbf{31.02}&\textbf{32.87}& 33.51 & 7.9$\times$  \\
 \bottomrule
\end{tabular}
}
% }
\caption{Results on WMT14 EN-DE/DE-EN and WMT16 EN-RO/RO-EN benchmarks. $I_{\text{dec}}$ is the number of decoding iterations and $m$ is the number of length reranking candidates. NPD represents noisy parallel decoding, CTC represents connectionist temporal classification. * indicate the results are obtained by our implementation.}% $N$ is the length of the output sequence.}
\label{tb.main_result}
\end{table*}

In this section, we first introduce the settings of our experiments, then report the main results compared with several strong baselines.
Ablation studies and further analysis are also included to verify the effects of different components used in \method.

\subsection{Experimental Settings}

\paragraph{Datasets}
We conduct experiments on three machine translation benchmarks: WMT14 EN-DE (4.5M translation pairs), WMT16 EN-RO (610k translation pairs), and IWSLT16 DE-EN (150K translation pairs).
These datasets are tokenized and segmented into subword units using BPE encodings~\citep{bpe}.
We preprocess WMT14 EN-DE by following the data preprocessing in \citet{transformer}. For WMT16 EN-RO and IWSLT16 DE-EN, we use the processed data provided in \citet{iter_nat}.

\paragraph{Knowledge Distillation}
Following previous work~\citep{nat,iter_nat,nat_reg}, we also use sequence-level knowledge distillation for all datasets. 
We employ the transformer with the base setting in ~\citet{transformer} as the teacher for knowledge distillation. Then, we train our \method on distilled data.

\paragraph{Baselines and Setup}
We compare our method with the base Transformer and strong representative NAT baselines in Table~\ref{tb.main_result}.
% including fully non-iterative models:  a vanilla NAT-base model, the NAT with fertility~\citep[NAT-FT]{nat}, the NAT imitating AT~\citep[imit-NAT]{imitate_nat}, the Flow-based NAT~\citep[Flowseq]{flowseq}, the NAT with hint-based training~\citep[NAT-HINT]{hint_nat}, Imputer~\citep{imputer}, the NAT with CRF~\citep{crf} decoding~\citep[NAT-DCRF]{nat-crf}, and the NAT with iterative refinement: NAR-IR~\citep{iter_nat}, LevT~\citep{levT}, Mask-Predict~\citep{mask_predict}, and JM-NAT~\citep{jm_nat}.
For all our tasks, we obtain other NAT models' performance by directly using the performance figures reported in their papers if they are available.

We adopt the vanilla model which copies source input uniformly in~\citet{nat} as our base model~(NAT-base) and replace the \textit{UniformCopy} with attention mechanism using positions. 
Note that the output length does not equal the length of reference in models using CTC. Therefore, for GLAT with CTC, we adopt longest common subsequence distance for comparing $Y$ and $\hat{Y}$, and the glancing target is the target alignment that maximize the output probability $arg\max_{a\in \mathcal{B}^{-1} (Y)} P(a|X;\theta)$. $\mathcal{B}^{-1}$ is the mapping proposed in \citep{ctc}, which expand the reference to the length of output by inserting blanks or repeating words.

For WMT datasets, we follow the hyperparameters of the base Transformer in \citet{transformer}. And we choose a smaller setting for IWSLT16, as IWSLT16 is a smaller dataset. For IWSLT16, we use 5 layers for encoder and decoder, and set the model size $d_{\text{model}}$ to 256.
Using Nvidia V100 GPUs, We train the model with batches of 64k/8k tokens for WMT/IWSLT datasets, respectively. We set the dropout rate to 0.1 and use Adam optimizer~\citep{adam} with $\beta=(0.9,0.999)$.
For WMT datasets, the learning rate warms up to $5e-4$ in 4k steps and gradually decays according to inverse square root schedule in~\citet{transformer}. As for IWSLT16 DE-EN, we adopt linear annealing (from $3e-4$ to $1e-5$) as in \citet{iter_nat}. For the hyper-parameter $\lambda$, we adopt linear annealing from 0.5 to 0.3 for WMT datasets and a fixed value of 0.5 for IWSLT16.
The final model is created by averaging the 5 best checkpoints chosen by validation BLEU scores.

\begin{figure}[!tbp]
\centering
\small
% \begin{minipage}[t]{0.45\textwidth}
\centering
\includegraphics[width=0.8\linewidth]{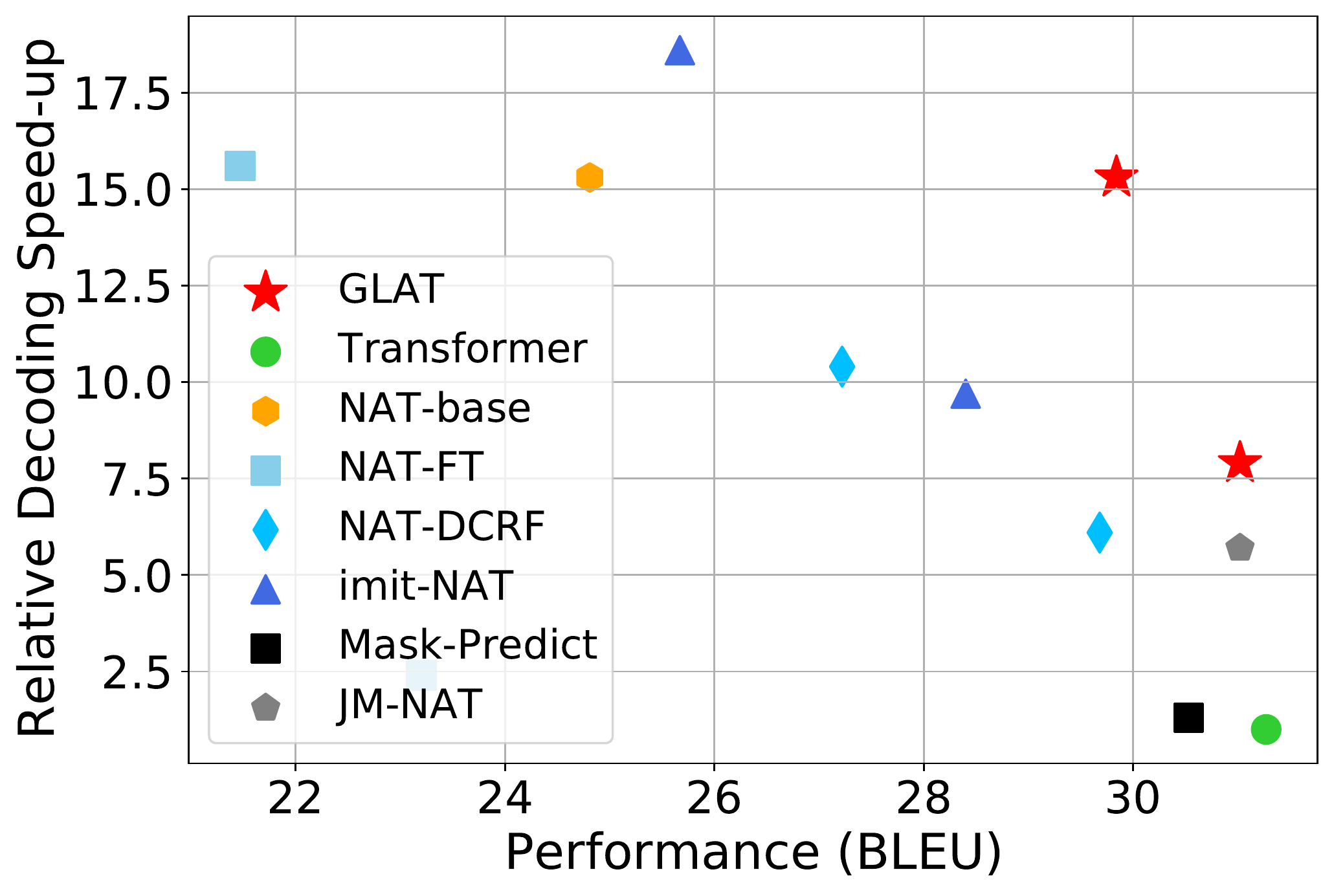}
\caption{The trade-off between speed-up and BLEU on WMT14 DE-EN}
\label{fig.trade_off}
% \end{minipage}
% \hspace{10pt}
\end{figure}

\subsection{Main Results}
The main results on the benchmarks are presented in Table~\ref{tb.main_result}.
\method significantly improves the translation quality and outperforms strong baselines by a large margin.
Our method introduces explicit word interdependency modeling for the decoder and gradually learns simultaneous generation of whole sequences, enabling the model to better capture the underlying data structure. 
%Our method introduces explicit target language modeling for the decoder and gradually learns simultaneous generation, which addresses \deadlock and better capture the underlying data structure.
%It is worth noting that although models equipped with iterative decoding achieve competitive BLEU~\citep{bleu} scores, their success is based on the sacrifice of speed advantage.
Compared to models with iterative decoding, our method completely maintains the inference efficiency advantage of fully non-autoregressive models, since \method generate with a single pass.
Compared with the baselines, we highlight our empirical advantages:
\begin{itemize}
    \item \method is highly effective. Compared with the vanilla NAT-base models, \method obtains significant improvements~(about 5 BLEU) on EN-DE/DE-EN. Additionally, \method also outperforms other fully non-autoregressive models with a substantial margin (almost +2 BLEU score on average). The results are even very close to those of the AT model, which shows great potential.
    
    %\item \method is  flexible, which is model-agnostic and can be applied to boost other NAT models.
    \item \method is simple and can be applied to other NAT models flexibly, as we only modify the training process by reference glancing while keeping inference unchanged. For comparison, NAT-DCRF utilizes CRF to generate sequentially; NAT-IR and Mask-Predict models need multiple decoding iterations.
    
    \item CTC and NPD use different approaches to determine the best output length, and they have their own advantages and disadvantages. CTC requires the output length to be longer than the exact target length. With longer output lengths, the training will consume more time and GPU memory. As for NPD, with a certain number of length reranking candidates, the inference speed will be slower than models using CTC. Note that NPD can use pretrained AT models or the non-autoregressive model itself to rerank multiple outputs.
    % Imputer and GLAT use different methods to determine the best target length. Based on CTC~\citep{ctc}, Imputer sets the max target length twice the length of the source input and determines the best length by removing blanks and contiguous repetitive words after generation. Thus,  it is non-trivial to apply target length reranking in Imputer, while GLAT can be further improved from 25.2 to 26.5 with AT reranking on WMT14 EN-DE, which outperforms the Imputer model. 
\end{itemize}

We also present a scatter plot in Figure~\ref{fig.trade_off}, displaying the trend of speed-up and BLEU with different NAT models.
It is shown that the point of \method is located on the top-right of the competing methods.
Obviously, \method outperforms our competitors in BLEU if speed-up is controlled, and in speed-up if BLEU is controlled.
This indicates that \method outperforms previous state-of-the-art NAT methods.
Although iterative models like Mask-Predict achieves competitive BLEU scores, they only maintain minor speed advantages over AT. In contrast, fully non-autoregressive models remarkably improve the inference speed.

\begin{figure}[!tbp]
%\begin{minipage}[t]{0.45\linewidth}
\centering
\includegraphics[width=0.8\linewidth]{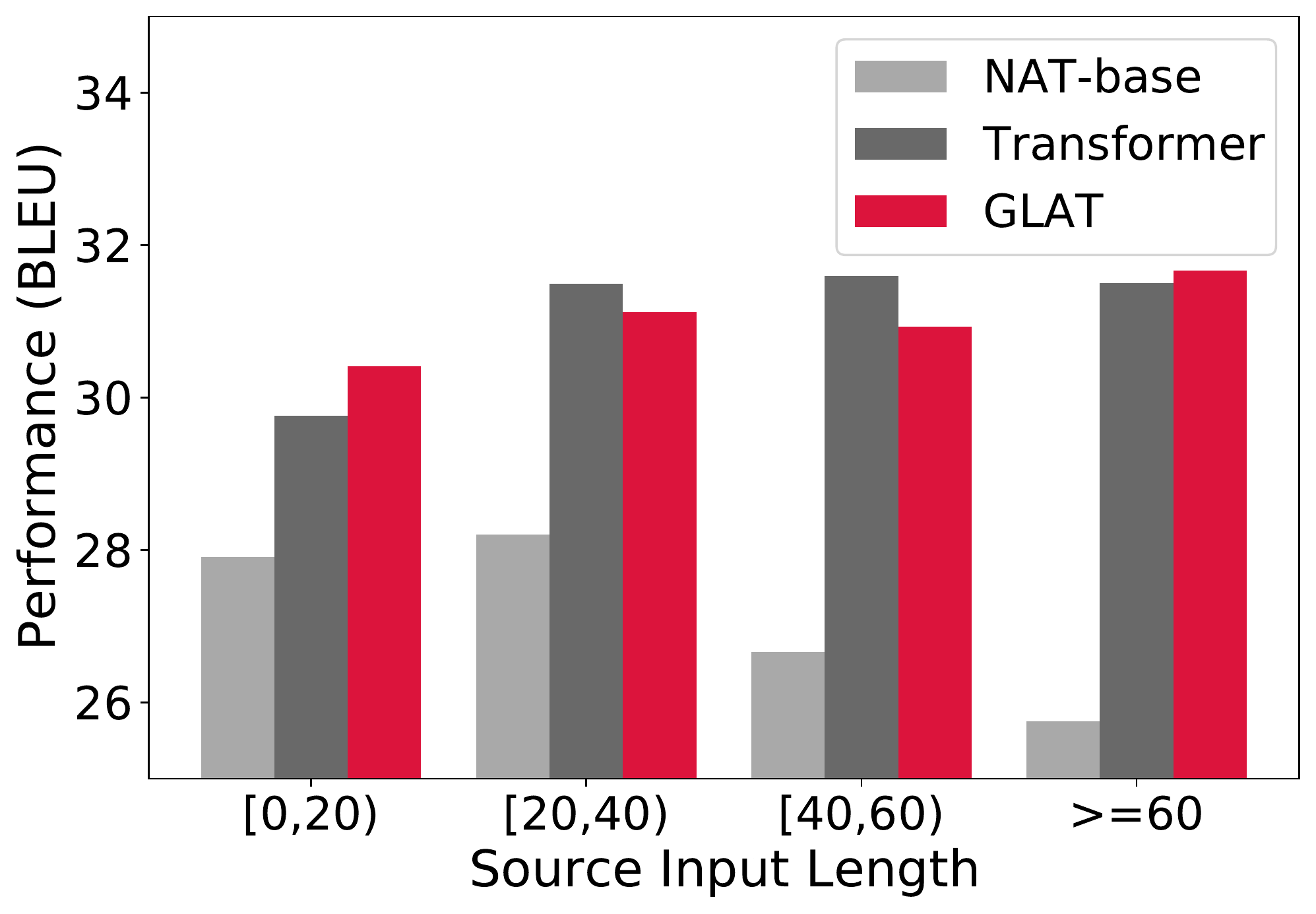}
\caption{Performance under different source input length on WMT14 DE-EN}
\label{fig.len_res}
%\end{minipage}
\end{figure}
\subsection{Analysis}
% \label{sec:analysis}
% \begin{figure}[tbp]
% \centering
% \small
% \begin{minipage}[t]{0.45\textwidth}
% \centering
% \includegraphics[width=0.8\linewidth]{figure/dec_iter3.pdf}
% \caption{The BLEU scores of \method with different decoding iterations}%\\ \qlh{need more comparison}}
% \label{fig.dec_iter}
% \end{minipage}
% \hspace{10pt}
% \begin{minipage}[t]{0.45\textwidth}
% \centering
% \includegraphics[width=0.8\linewidth]{figure/freeze_enc.pdf}
% \caption{Training NAT with different initialized encoder}
% \label{fig.freeze_enc}
% \end{minipage}
% \end{figure}

\paragraph{Effect of Source Input Length}
To analyze the effect of source input length on the models' performance, we split the source sentences into different intervals by length after BPE and compute the BLEU score for each interval. 
The histogram of results is presented in Figure~\ref{fig.len_res}. 
NAT-base's performance drops sharply for long sentences, while the gradual learning process enables GLAT to boost the performance by a large margin, especially for long sentences. We also find that GLAT outperforms autoregressive Transformer when the source input length is smaller than 20.

\paragraph{GLAT Reduces Repetition}
We also measure the percentage of repeated tokens on test set of WMT14 EN-DE and WMT14 DE-EN. Table~\ref{tb.repetition} presents the token repetition ratio of sentences generated by NAT-base and GLAT.
The results show that GLAT significantly reduces the occurrence of repetition, and the repetition ratio can be further reduced with NPD. We think an important cause of the improvement is better interdependency modeling. Since GLAT explicitly encourages word interdependency modeling to better capture the dependency between target tokens, wrong generation patterns, such as repetition, can be largely avoided.

\begin{table}[!tbp]
\centering
\small
% \scalebox{0.9}{
\begin{tabular}{lcc}
\toprule
\multirow{2}{*}{Model} & \multicolumn{2}{c}{WMT14} \\
& EN-DE & DE-EN \\
\midrule
NAT-base & 8.32\% & 7.10\%\\
GLAT & 1.19\% & 1.05\% \\
GLAT w/ NPD & 0.32\% & 0.16\% \\
\bottomrule
\end{tabular}
\caption{Token repetition ratio on WMT14 EN-DE and WMT14 DE-EN}
\label{tb.repetition}
\end{table}

\subsection{Ablation Study}

\paragraph{Effectiveness of the Adaptive Sampling Number}
To validate the effectiveness of the adaptive sampling strategy for the sampling number $S(Y,\hat Y)$, we also introduce two fixed approaches for comparison. The first one decides the sampling number with $\lambda*T$, where $T$ is the length of $Y$, and $\lambda$ is a constant ratio. The second one is relatively flexible,  which sets a start ratio of $\lambda_s$ and an end ratio $\lambda_e$, and linearly reduces the sampling number from $\lambda_s*T$ to $\lambda_e*T$ along the training process.

As shown in Table~\ref{tb.constant_base} and Table~\ref{tb.adaptive_base}, clearly, our adaptive approach~(Adaptive in the table) outperforms the baseline models with big margins. The results confirm our intuition that the sampling schedule affects the generation performance of our NAT model.
The sampling strategy, which first offers relatively easy generation problems and then turns harder, benefits the final performance.
Besides, even with the simplest constant ratio, \method still achieves remarkable results. When set $\lambda=0.2$, it even outperforms the baseline $\lambda=0.0$ by 2.5 BLEU score.

The experiments potentially support that it is beneficial to learn the generation of fragments at the start and gradually transfer to the whole sequence. The flexible decreasing ratio method works better than the constant one, and our proposed adaptive approaches achieve the best results.

\begin{table}[!tbp]
\centering
\small
\begin{tabular}{lcc}
\toprule
Sampling Number &  $\lambda$  & BLEU \\
\midrule
\multirow{5}{*}{Fixed}  & 0.0  & 24.66  \\
                        & 0.1  & 24.91 \\
                        & 0.2  & 27.12  \\
                        & 0.3  & 24.98  \\
                        & 0.4  & 22.96  \\
\midrule
Adaptive                & -  & \textbf{29.61}  \\
\bottomrule
\end{tabular}
\caption{Performances on IWSLT16 with fixed sampling ratio.}
\label{tb.constant_base}
\end{table}

\begin{table}[!tbp]
\small
\centering
\begin{tabular}{lccc}
\toprule
Sampling Number& $\lambda_s$ & $\lambda_e$ & BLEU \\
\midrule
\multirow{3}{*}{Decreasing} & 0.5 & 0 & 27.80  \\
& 0.5 & 0.1 & 28.21  \\
& 0.5 & 0.2 & 27.15 \\
& 0.5 & 0.3 & 23.37 \\
\midrule
Adaptive & \multicolumn{2}{c}{-} & \textbf{29.61}  \\
\bottomrule
\end{tabular}
\caption{\label{tb.adaptive_base} Performances on IWSLT16 with decreasing sampling ratio.}
\end{table}
\paragraph{Influence of Reference Word Selection}
To analyze how the strategies of selecting reference words affect glancing sampling, we conduct experiments with different selection strategies. By default, we assume all the words in the reference are equally important and randomly choose reference words for glancing.
%that is, we assume the probability of each word in the reference is equal and sample from a uniform distribution.
Besides the random strategy, we devise four other selection methods considering the prediction of first decoding. 
For $p_\text{ref}$ and $1-p_\text{ref}$, the sampling probability of each reference word is proportional to the output probability for the reference word $p_\text{ref}$ and the probability $1-p_\text{ref}$, respectively. Similar to the word selection strategy for masking words during inference in Mask-Predict, we also add two strategies related to the prediction confidence: "most certain" and "most uncertain." We choose the positions where predictions have higher confidence for "most certain", and vise versa for "most uncertain." The results for different selection methods are listed in Table~\ref{tb.sp_strategy}.

In comparisons, the model with the selection strategy $1-p_\text{ref}$ outperforms the one with $p_\text{ref}$, indicating that words hard to predict are more important for glancing in training. 
And we find that the random strategy performs a little better than the two confidence-based strategies. We think this indicates that introducing more randomness in sampling enable \method to explore more interdependency among target words. We adopt the random strategy for its simplicity and good performance.

\begin{table}[!t]
\centering
\small
\tabcolsep 4pt
\begin{tabular}{lcc}
\toprule
Selection Strategy & \method & \method w/ NPD \\
\midrule
random             & 25.21  & 26.55        \\
$p_\text{ref}$            & 24.87  & 25.83        \\
$1-p_\text{ref}$          & 25.37  & 26.52        \\
most certain              & 24.99  & 26.22        \\
most uncertain            & 24.86  & 26.13       \\
\bottomrule
\end{tabular}
\caption{Performance on WMT14 EN-DE with different reference word selection strategies.}
\label{tb.sp_strategy}
\end{table}

\begin{table}[!tp]
\centering
\small
\tabcolsep 4pt
\begin{tabular}{l|cc}
\toprule
\multirow{2}{*}{Method}& \multicolumn{2}{c}{WMT14} \\
& EN-DE & DE-EN\\
\midrule
\method w/ uniform sampling &  19.16 &  23.56 \\
\method w/ \texttt{[MASK]} inputs & 24.99 & 29.48 \\
\method & 25.21 & 29.84  \\
\bottomrule
\end{tabular}
\caption{\label{tb.ablation} Ablation study for comparing GLAT and Mask-Predict on WMT14 EN-DE and DE-EN.}
\end{table}
% In comparison, the model with sampling distribution where falsely predicted words have higher probability achieves better performance than the one which increases probabilities for correctly predicted words, indicating that words hard to predict are more important for glancing in the training process.
% Besides, we find that the performance of random selection is similar to $1-p_\text{ref}$ and the two selection methods based on prediction certainty perform a little worse.

% \begin{table*}[!tp]
% \centering
% \tabcolsep 4pt
% \small
% \begin{tabular}{l|cccc}
% \toprule
% \multirow{2}{*}{Method}& \multicolumn{2}{c}{WMT14 EN-DE} &\multicolumn{2}{c}{WMT14 DE-EN}\\
% & Hamming  & Levenshtein & Hamming & Levenshtein \\
% \midrule
% \method & 25.21 & 24.56 & 29.84 & 28.96 \\
% \method(w/ reranking m=7) & 26.55 & 26.21 & 31.02 & 30.85 \\
% \bottomrule
% \end{tabular}
% \caption{\label{tb.distance} Performance on WMT14 EN-DE and WMT14 DE-EN with different distances.}
% \end{table*}

\paragraph{Advantages of \method over Mask-Predict}
To study the effects of sampling strategy and decoder inputs of \method, we conduct experiments for replacing these two modules in \method with the corresponding part in Mask-Predict, respectively. 
The results are presented in Table~\ref{tb.ablation}. \method employs glancing sampling strategy instead of the uniform sampling strategy used in Mask-Predict, and replaces the \texttt{[MASK]} token inputs with source representations from the encoder. The results show that the glancing sampling strategy outperforms the uniform sampling strategy by 5$\sim$6 BLEU points, and feeding representations from the encoder as the decoder input could still improve the strong baseline by 0.2$\sim$0.3 BLEU points after adopting glancing sampling.
To sum up, the adaptive glancing sampling approach contributes the most to the final improvement, and the use of representations from the encoder also helps a bit.

\paragraph{More Analysis}
We also conduct experiments for: a) comparison of different distance metrics for glancing sampling, b) \method with more than one decoding iteration. The details are in Appendix.

%% file: 050related.tex
%Since~\citet{nat} proposed the non-autoregressive transformer~(NAT) which enables parallel sequence generation, there have been several advances in developing non-autoregressive models.

\paragraph{Fully Non-Autoregressive Models} A line of work introduces various forms of latent variables to reduce the model's burden of dealing with dependencies among output words ~\citep{nat,flowseq,pnat,reorder_nat}.
%~\citet{nat} interprets the latent variable as the number of target words aligned to each source words. \citet{flowseq} utilized the generative flow to model expressive signals related to the output. ~\citet{pnat} modeled the position of output words, to address the word reordering issue. ~\citet{reorder_nat} reordered the input sequence as the intermediate translation of output sequence. 
Another branch of work considers transferring the knowledge from autoregressive models to non-autoregressive models~\citep{imitate_nat,hint_nat,nat_fcl,nat_em}.
Besides, there are also some work that apply different training objectives to train non-autoregressive models~\citep{nat_ctc,nat_bon,nat_axe}, add regularization terms~\citep{nat_reg,enat}.
% ~\citet{imitate_nat} guided the training via imitating demonstrations from modules in autoregressive models.
% ~\citet{hint_nat} leveraged the relevance between hidden states and the attention distribution of autoregressive models.
%Compared to fully non-autoregressive models, our proposed method stays simple and can significantly boost the performance without the need of modifying model architectures or the inference process.
%Our proposed method directly introduce explicit target language modeling into NAT, rather than transferring knowledge from the AT model.
\paragraph{Non-Autoregressive Models with Structured Decoding} To model the dependencies between words, \citet{nat-crf} introduces a CRF inference module in NAT and performs additional sequential decoding after the non-autoregressive computation in inference. \citet{cascade_decoding} proposes cascaded CRF decoding. Since \method only performs single-pass non-autoregressive generation, our approach is orthogonal to the method proposed in~\citet{nat-crf}.  We can also combine our approach with the structured decoding methods.

\paragraph{Non-Autoregressive Models with Iterative Refinement} A series of work are devoted to semi-autoregressive models that  refine the outputs with multi-pass iterative decoding~\citep{iter_nat,levT,mask_predict,smart, disco}.
~\citet{iter_nat} proposed a method of iterative refinement based on denoising autoencoder.
~\citet{levT} utilized insertion and deletion to refine the outputs in inference.
~\citet{mask_predict} trained the model with the masked language model, and the model iteratively replaces masked tokens with new outputs.
Despite the relatively better accuracy, the multiple decoding iterations vastly reduce the inference efficiency of non-autoregressive models.

%% file: 060conclusion.tex
%In non-autoregressive models, learning is under strong conditional independence assumption and lacks explicit target language modeling, which brings a challenge for training.
In this paper, we propose Glancing Transformer with a glancing language model to improve the performance of single-pass parallel generation models. 
With the glancing language model, the model starts from learning the generation of sequence fragments and gradually moving to whole sequences. 
Experimental results show that our approach significantly improves the performance of non-autoregressive machine translation with single-pass parallel generation. As \method achieves competitive performance compared with autoregressive models, applying our approach to other generation tasks is a promising direction for future work.

% As non-autoregressive models are efficient and have great potential in multiple tasks, we plan to apply our approach to other tasks.